\documentclass{article}

\PassOptionsToPackage{numbers, sort&compress}{natbib}

\usepackage[final]{neurips_2024}

\usepackage[utf8]{inputenc} 
\usepackage[T1]{fontenc}    
\usepackage[colorlinks=true, linkcolor=red, urlcolor=blue, citecolor=green]{hyperref}       
\usepackage{url}            
\usepackage{booktabs}       
\usepackage{amsfonts}       
\usepackage{nicefrac}       
\usepackage{microtype}      
\usepackage{xcolor}         

\usepackage{wrapfig}
\usepackage{graphicx}
\usepackage{bm}
\newcommand{\E}{\operatorname{\mathbb E}}
\newcommand{\R}{\operatorname{\mathbb R}}
\usepackage{amsmath}
\usepackage{subcaption}
\usepackage{kotex}

\definecolor{mygreen}{RGB}{112, 173, 71}
\definecolor{myorange}{RGB}{237, 125, 49}
\definecolor{myblue}{RGB}{29, 73, 177}
\usepackage[capitalize,noabbrev]{cleveref}

\title{Diversify, Contextualize, and Adapt:\\ Efficient Entropy Modeling for Neural Image Codec}

\author{%
  Jun-Hyuk Kim\thanks{\footnotesize Corresponding author} \quad Seungeon Kim \quad Won-Hee Lee \quad Dokwan Oh \\
  Samsung Advanced Institute of Technology \\
  \texttt{\{jh131.kim, se2.kim, why\_wh.lee, dokwan.oh\}@samsung.com} \\
}

\begin{document}

\maketitle

\begin{abstract}
    Designing a fast and effective entropy model is challenging but essential for practical application of neural codecs. Beyond spatial autoregressive entropy models, more efficient backward adaptation-based entropy models have been recently developed. They not only reduce decoding time by using smaller number of modeling steps but also maintain or even improve rate--distortion performance by leveraging more diverse contexts for backward adaptation. Despite their significant progress, we argue that their performance has been limited by the simple adoption of the design convention for forward adaptation: using only a single type of hyper latent representation, which does not provide sufficient contextual information, especially in the first modeling step. In this paper, we propose a simple yet effective entropy modeling framework that leverages sufficient contexts for forward adaptation without compromising on bit-rate. Specifically, we introduce a strategy of diversifying hyper latent representations for forward adaptation, i.e., using two additional types of contexts along with the existing single type of context. In addition, we present a method to effectively use the diverse contexts for contextualizing the current elements to be encoded/decoded. By addressing the limitation of the previous approach, our proposed framework leads to significant performance improvements. Experimental results on popular datasets show that our proposed framework consistently improves rate--distortion performance across various bit-rate regions, e.g., 3.73\% BD-rate gain over the state-of-the-art baseline on the Kodak dataset.      
\end{abstract}

\section{Introduction}
\label{sec:introduction}

Most neural image codecs~\cite{minnen2018joint,minnen2020channel,he2021checkerboard,he2022elic,kim2022joint,liu2023learned} first transform an image into a quantized latent representation.
It is then encoded into a bitstream via an entropy coding algorithm, which relies on a learned probability model known as the entropy model.
According to the Shannon's source coding theorem, the minimum expected length of a bitstream is equal to the entropy of the source.
Thus, accurately modeling entropy of the quantized latent representation is crucial.

Entropy models estimate a joint probability distribution over the elements of the quantized latent representation.
Generally, it is assumed that all elements follow conditionally independent probability distributions.
To satisfy this, the probability distributions are modeled in context-adaptive manners, which is key to accurate entropy modeling~\cite{minnen2018joint}. 
Recent methods are based on the joint backward and forward adaptation where the probability distributions adapt by leveraging contexts in two different ways: directly using previously encoded/decoded elements (i.e., backward adaptation), and extracting and utilizing an additional hyper latent representation (i.e., forward adaptation).
Here, the type of contexts leveraged can be diverse depending on the spatial range they cover. 
First, each element has dependencies with other elements in the same spatial location along the channel dimension.
Since the channel-wise dependencies correspond to the local image area (e.g., a $16\times16$ patch), we denote them as the ``local'' context.
Second, dependencies exist among spatially adjacent elements, and we refer to them as the ``regional'' context. 
Lastly, long-range spatial dependencies span the entire image area, referred to as the ``global'' context. 

For the backward adaptation, the modeling order, i.e., which elements are modeled first, is an important factor, and the key lies in how effectively we can utilize diverse contexts in the modeling process.
Early studies employ spatial autoregressive (AR) models that access regional context including the most spatially adjacent elements. However, they suffer from significantly slow decoding times due to the inevitably large number of modeling steps, which is equal to the spatial dimensions~\cite{minnen2018joint}.
To enhance efficiency in entropy modeling, several attempts reduce the number of modeling steps while leveraging diverse contexts: a 10-step channel-wise AR model~\cite{minnen2020channel}, a 2-step spatial non-AR model with a checkerboard pattern~\cite{he2021checkerboard}, and a 4-step non-AR model that operates across spatial and channel dimensions using a quadtree partition~\cite{li2023dcvc}.

\begin{wrapfigure}[25]{r}{0.5\textwidth}
\vspace{-1.5em}
\centering
\includegraphics[width=.98\linewidth]{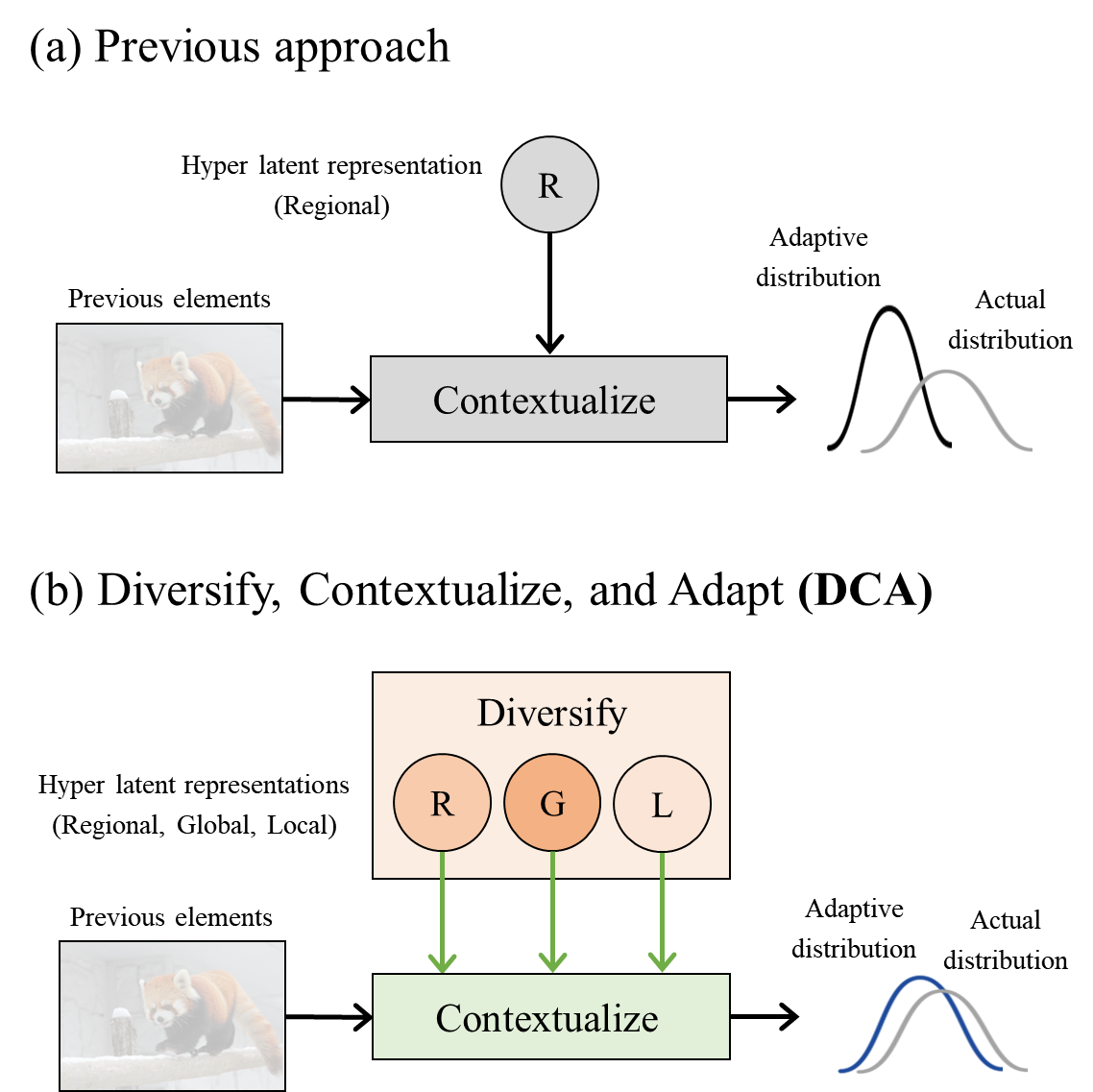}
\caption{\textbf{DCA} \textcolor{myorange}{\textbf{diversifies}} the hyper latent representations and \textcolor{mygreen}{\textbf{contextualizes}} the current elements by leveraging the diverse hyper latent representations along with the previous elements. As a result, the probability distributions \textcolor{myblue}{\textbf{adapt}} effectively, leading to accurate entropy modeling.}
\label{fig:teaser}
\end{wrapfigure}

Entropy models based on the efficient backward adaptation methods have led to significant improvements.
However, they are still limited in fully leveraging contexts for forward adaptation.
Since they use multiple neural layers with downsampling and upsampling for modeling hyper latent representation, they can only access the regional context. 
This limits the performance improvement due to the insufficient contexts (\cref{fig:teaser}a).
In particular, this limitation is exacerbated at the first step where only forward adaptation is utilized due to the absence of previous elements (\cref{fig:latent_analysis_1}). 
Therefore, it is necessary to develop effective forward adaptation in synergy with the efficient backward adaptation.

In this paper, we propose a simple yet effective entropy modeling framework, called \textbf{DCA} (\underline{D}iversify, \underline{C}ontextualize, and \underline{A}dapt), leveraging sufficient contexts for forward adaptation without compromising on bit-rate (\cref{fig:teaser}b).
Building on the quadtree partition-based backward adaptation~\cite{li2023dcvc}, we introduce a strategy of diversification, i.e., extracting local, regional, and global hyper latent representations unlike only a single regional one in the previous approach.
Note that simply using more contexts for forward adaptation does not guarantee performance improvements because forward adaptation requires additional bit allocation unlike backward adaptation.
Then, we propose how to effectively utilize the diverse contexts along with the previously modeled elements for contextualizing the current elements to be encoded/decoded.
To consider step-wise different situations, e.g., increased number of previous elements over steps, our contextualization method is designed to utilize each hyper latent representation separately in a step-adaptive manner.
Additionally, our contextualization method proceeds in the sequence of regional, global, and local hyper latent representations.
Similarly to backward adaptation, we empirically observe that modeling order also matters in forward adaptation.  

Our main contributions are summarized as follows:
\begin{itemize}
    \item We propose a strategy of diversifying contexts for forward adaptation by extracting three different hyper latent representations, i.e., local, regional, and global ones. This strategy can provide sufficient contexts for forward adaptation without compromising on bit-rate.    
    \item We introduce how to effectively leverage the diverse contexts, i.e., previously modeled elements and the three hyper latent representations. We empirically show that the modeling order of three types of contexts affects the performance.   
    \item Through the diversification and contextualization methods, our DCA effectively adapts, resulting in significant performance improvements. For example, DCA achieves 3.73\% BD-rate gain over the state-of-the-art method~\cite{li2023dcvc} on the Kodak dataset.      
\end{itemize}

\section{Related work}
\label{sec:related_work}

\paragraph{Joint backward and forward adaptation.}
\citet{balle2018variational} propose a scale hyperprior for forward adaptation.
A hyper latent representation is extracted and utilized for inferring local scale parameters of the parameterized entropy model.
\citet{minnen2018joint} extend the hyperprior model by using an additional mean hyperprior, and introduce joint backward and forward adaptation by combining the extended hyperprior model with a spatial autoregressive (AR) model.
A patch matching-based non-local referring model \cite{qian2021global} and a multi-head attention-based global hyperprior \cite{kim2022joint} are proposed to enrich contexts for backward and forward adaptation, respectively. 

\paragraph{Efficient backward adaptation.}
To address the slow decoding times of spatial AR-based entropy models, several studies have proposed group-wise backward adaptation methods.
They first divide the quantized latent representation into multiple groups and then process them in a group-wise manner, resulting in improved efficiency.
\citet{he2021checkerboard} propose dividing the quantized latent representation into two groups using the checkerboard pattern, which is further improved by incorporating Transformer-based modules \cite{qian2022entroformer}.
While they apply a group-wise modeling in spatial dimension,
\citet{minnen2020channel} introduce a channel-wise AR model that divides the quantized latent representation into ten groups along channel dimension.
Some studies \cite{zhu2022transformer, zou2022devil} improve this model by applying Swin Transformer \cite{liu2021swin}.
Based on the channel-wise AR model, \citet{he2022elic} optimize the channel division and combine it with the checkerboard-based model. 
Recently, \citet{li2023dcvc} propose a quadtree partition-based backward adaptation that divides the quantized latent representation into four groups considering both channel and spatial dimensions.

In this paper, we propose a novel fast and effective entropy model that achieves better rate--distortion performance by diversifying not only the quantized latent representation but also the hyper latent representations for backward and forward adaptation, respectively.
\section{Methods}
\label{sec:methods}

\begin{figure}[t]
\begin{center}
\centerline{\includegraphics[width=\linewidth]{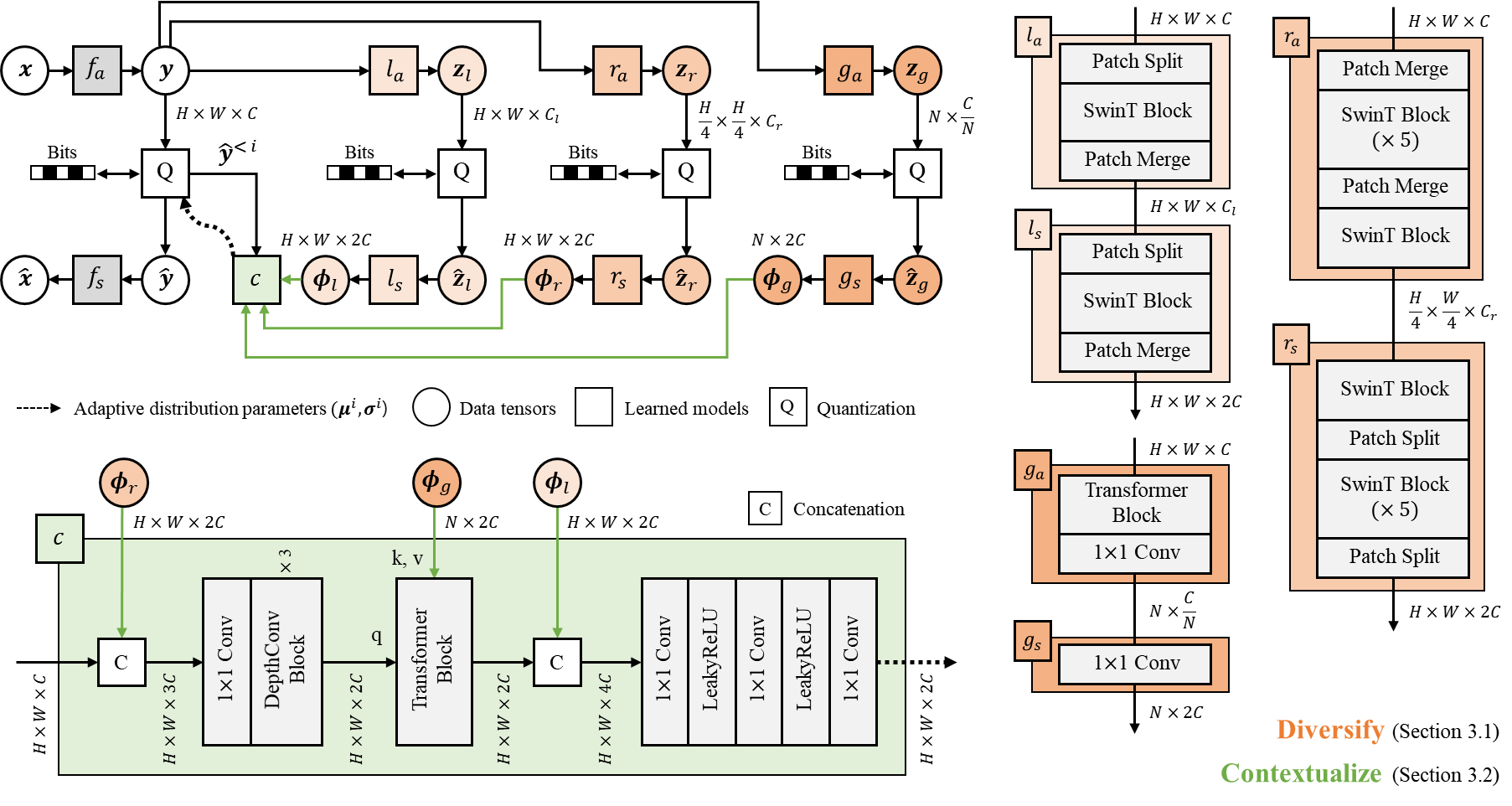}}
\caption{
Overview of the neural image codec with the proposed entropy model, referred to as DCA.
DCA can be employed by any analysis and synthesis transforms $f_a(\cdot)$ and $f_s(\cdot)$.
DCA is an adaptive entropy model consisting of two main stages: \textcolor{myorange}{\textbf{diversify}} (\cref{subsec:diversify}) and \textcolor{mygreen} {\textbf{contextualize}} (\cref{subsec:contextualize}).
First, given the latent representation $\bm y$, DCA extracts diverse hyper latent representations $\hat{\bm z}_l$, $\hat{\bm z}_r$, and $\hat{\bm z}_g$, and then encodes them into the bitstreams using learned factorized entropy models, which are omitted in this figure for simplicity.
Second, contextualization proceeds over four steps.
By using the three features $\bm \phi_l$, $\bm \phi_r$, and $\bm \phi_g$ (from the three hyper latent representations, respectively) and all the previously encoded/decoded elements before the $i$-th step, i.e., $\hat{\bm y}^{<i}$, DCA contextualizes the current elements to be encoded/decoded, i.e., $\hat{\bm y}^{i}$, and finally obtains adaptive distribution parameters $\bm{\mu}^i$ and $\bm{\sigma}^i$ for probability modeling.
Using the learned adaptive probability model, the quantized latent representation $\hat{\bm y}$ are encoded into a bitstream.
}
\label{fig:overview}
\end{center}
\end{figure}

We provide an overview of the proposed methods in \cref{fig:overview}. 
The analysis transform $f_a(\cdot)$ and the synthesis transform $f_s(\cdot)$ (the gray blocks in \cref{fig:overview}) are learned to find an effective mapping between an input image $\bm x$ and a quantized latent representation $\hat{\bm y}$, i.e., $\hat{\bm y} = \lfloor{f_a(\bm x)}\rceil$ and $\hat{\bm x} = f_s(\hat{\bm y})$,
where $\left\lfloor{\cdot}\right\rceil$ is a round operation and $\hat{\bm x}$ is the decoded image.
For the analysis and synthesis transforms, we adopt the same model structure as in the ELIC-sm model~\cite{he2022elic} due to its efficiency.

All other components are learned to model a prior probability distribution on the quantized latent representation $\hat{\bm y}$, i.e., the entropy model $p_{\hat{\bm y}}$.
The learned entropy model is utilized in the process of entropy coding, for which we employ the asymmetric numeral systems~\cite{duda2013asymmetric}.
Here, our goal is to design a fast and effective learned entropy model.

\begin{wrapfigure}[18]{r}{0.5\textwidth}
\vspace{-1em}
\centering
\includegraphics[width=\linewidth]{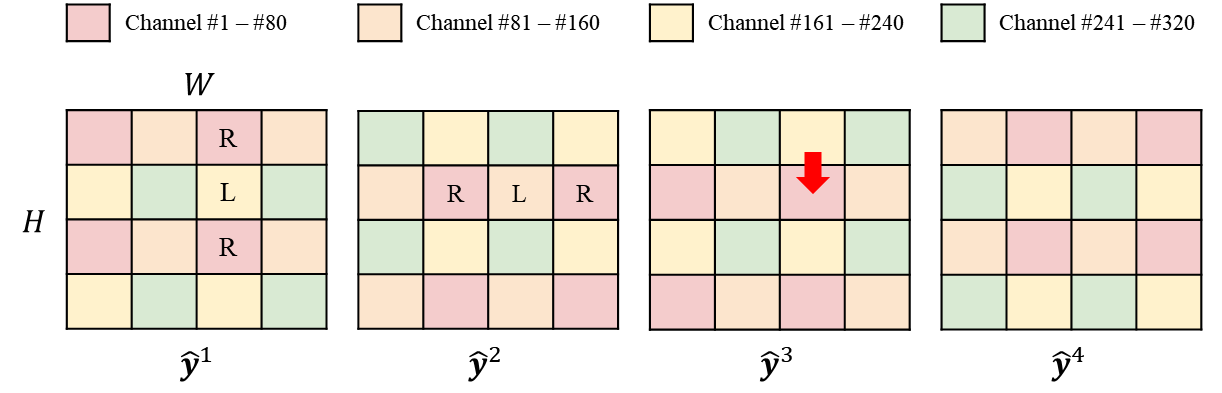}
\caption{Example of the quadtree partition-based backward adaptation for $\hat{\bm y}\in \R^{4\times 4\times 320}$.
For simplicity, channel dimensions are represented via  different colors.
$\hat{\bm y}^i$ means the elements to be encoded/decoded at the $i$-th step. 
For modeling the current elements $\hat{\bm y}^i$, all the previous modeled elements $\hat{\bm y}^{< i}$ are used.
For example, the elements corresponding to the red arrow leverage diverse contexts including elements across different channels at the same spatial location (local context denoted as L) and spatially adjacent four elements of the same channel (regional context denoted as R).}
\label{fig:quadtree}
\end{wrapfigure}

\paragraph{Quadtree partition.}
We build our entropy model on the joint forward and backward adaptation where the quadtree partition is used, which is formulated as follows \cite{li2023dcvc}: 
$$p(\hat{\bm{y}}) = p(\hat{\bm{z}}) \times p(\hat{\bm{y}} | \hat{\bm{z}}) = p(\hat{\bm{z}}) \times \prod_{i=1}^{4} p(\hat{\bm{y}}^i | \hat{\bm{y}}^{<i}, \hat{\bm{z}})$$
where $\hat{\bm y}$ is the quantized latent representation, $\hat{\bm{z}}$ is the quantized regional hyper latent representation, $\hat{\bm y}^{i}$ is the elements to be modeled at the $i$-th step, and $\hat{\bm y}^{<i}$ is all the previous modeled elements before the $i$-th step. At each step, one-fourth of the total elements are modeled.
The method partitions the quantized latent representation into four groups along the channel dimension, and then divides each group into non-overlapping 2$\times$2 patches along the spatial dimension. The entropy modeling proceeds over four steps, with each step modeling different elements as shown in \cref{fig:quadtree}. 
This quadtree partition-based method uses diverse contexts for backward adaptation, capturing dependencies from both spatial and channel dimensions.

\paragraph{Motivation.}
Recent studies to efficient modeling of backward adaptation have made significant advancements in terms of optimizing the rate--distortion--computation trade-off; however, there is still a gap between their assumptions in the probability modeling and actual data, leaving room for further performance enhancement.
The assumptions are as follows: 1) All elements of $\hat{\bm{z}}$ are independent; 2) All elements of $\hat{\bm{y}}^i$ are conditionally independent given $\hat{\bm{y}}^{<i}$ and $\hat{\bm{z}}$.
Here, the more the actual data deviates from the assumptions, the lower the accuracy of the entropy modeling. 
At the first modeling step, the elements are modeled conditioned only on the quantized hyper latent representation, i.e., $p(\hat{\bm y}^1 | \hat{\bm{z}})$, resulting in a hyperprior model known for deviating from to the second assumption~\cite{minnen2018joint}.
This can be more problematic because the state-of-the-art methods process a relatively large number of elements at the first step in order to complete the overall modeling with a minimal number of steps. We also empirically show that this problem actually occurs in \cref{fig:latent_analysis_1}.

One straightforward solution is to increase the number of steps so that fewer elements are modeled in the first step. However, this leads to slower modeling speeds, which conflict with the goal of our paper, i.e., developing a fast and effective entropy model. 
Another simple approach is to provide more quantized regional hyper latent representation $\hat{\bm{z}}$ when modeling the quantized latent representation $\hat{\bm{y}}^{1}$. However, paradoxically, this approach can introduce another issue due to the first assumption. 
Since all elements of hyper latent representation are the same type of information (i.e., regional context), there is a relatively high likelihood of dependencies among the elements. 
Therefore, to meet both assumptions, the newly added hyper latent representation is required to be independent from the existing regional hyper latent representation. This is why our proposed diversification method using local, regional, and global hyper latent representations is needed. 
In this paper, we propose a fast and effective entropy model, called DCA, which consists of three main stages: diversifying the hyper latent representations, contextualizing the elements targeted for probability modeling, and ultimately adapting the probability distribution of the elements to the given contexts.

\subsection{Diversify}
\label{subsec:diversify}

The proposed DCA aims to diversify the information that the hyper latent representations contain.
Specifically, given the latent representation $\bm y \in \R^{H\times W\times C}$, where $H$, $W$, and $C$ are the height, width, and the number of channels, respectively, DCA extracts three different types of hyper latent representations depending on the range they cover: a local hyper latent representation $\hat{\bm z}_l \in \R^{H\times W\times C_l}$, a regional hyper latent representation $\hat{\bm z}_r \in \R^{\frac{H}{4}\times \frac{W}{4}\times C_r}$, and a global hyper latent representation $\hat{\bm z}_g \in \R^{N\times \frac{C}{N}}$.
The whole process is illustrated in the orange blocks of \cref{fig:overview}.

\paragraph{Local context.}
To model remaining dependencies along channel dimension at each spatial location, which correspond to a $16\times 16$ local patch in the image domain, we introduce local hyper analysis and synthesis transforms, $l_a(\cdot)$ and $l_s(\cdot)$, based on Swin Transformer (SwinT)~\cite{liu2021swin}.
The local hyper analysis transform $l_a(\cdot)$ analyzes local information in the latent representation, followed by the quantization operation to obtain a local hyper latent representation $\hat{\bm z}_l$.
The local synthesis transform $l_s(\cdot)$ synthesizes the local features $\bm \phi_l\in\R^{H\times W\times 2C}$ for contextualization from the local hyper latent representation $\hat{\bm z}_l$.

Each transform proceeds in the order of a Patch Split block, a SwinT block, and a Patch Merge block.
The Patch Split block serves the function of shifting all channel-wise elements at each spatial location to a $2\times2$ spatial resolution, consisting of the depth-to-space, layer normalization, and linear layers in sequence.
The SwinT block then captures dependencies between elements within each non-overlapping window of the input, producing an output of the same size as the input.
By setting the window size to $2\times 2$ in conjunction with the use of the Split block, we enforce the local hyper transforms to focus only on the local image area.
The Patch Merge block performs the opposite function of the Patch Split block, containing the layer normalization, linear, and space-to-depth layers in sequence.

\paragraph{Regional context.}
While the receptive field of the local hyper transforms is limited to the local image area (i.e., $16\times16$ patches), regional hyper analysis transform $r_a(\cdot)$ and regional hyper synthesis transform $r_s(\cdot)$ model remaining dependencies between elements distributed across a relatively wide image area.
The regional hyper analysis transform $r_a(\cdot)$ analyzes regional information in the latent representation and yields a regional hyper latent representation $\hat{\bm z}_r$ after quantization.
From the extracted regional hyper latent representation $\hat{\bm z}_r$, the regional synthesis transform $r_s(\cdot)$ generates the regional features $\bm \phi_r\in\R^{H\times W\times 2C}$ for contextualization.

To do this, we stack multiple layers with the downsampling and upsampling operations for the regional hyper analysis and synthesis transforms, respectively. 
We adopt the same structure as the previous work~\cite{zhu2022transformer}, which is based on SwinT.
Specifically, the regional hyper analysis transform $r_a(\cdot)$ conducts a Patch Merge block, five SwinT blocks, a Patch Merge block, and a SwinT block.
The regional hyper synthesis transform $r_s(\cdot)$ is constructed in the opposite order of the hyper analysis transform $r_a(\cdot)$, using the Patch Split block instead of the Patch Merge block.

\paragraph{Global context.}
Lastly, to capture remaining dependencies between elements across the whole image area, we construct global hyper analysis and synthesis transforms $g_a(\cdot)$ and $g_s(\cdot)$ by adopting model structure of the global hyperprior model of Informer~\cite{kim2022joint}.
The global hyper analysis transform $g_a(\cdot)$ extracts globally abstracted information from the latent representation using a Transformer block with cross-attention and a $1\times 1$ convolutional layer.
After quantization, it obtains a global hyper latent representation $\hat{\bm z}_g$. 
Using a $1\times 1$ convolutional layer, the global synthesis transform $g_s(\cdot)$ infers the global features $\bm \phi_g\in\R^{N\times 2C}$ for contextualization. 

\subsection{Contextualize}
\label{subsec:contextualize}
Diverse contexts, i.e., previously modeled elements (i.e., $\hat{\bm{y}}^{<i}$) and hyper latent representations (i.e., $\hat{\bm z}_l$, $\hat{\bm z}_r$, and $\hat{\bm z}_g$) can be used for adapting probability distributions.
Here, an important research question emerges: How can we effectively leverage the diverse contexts?
First, to consider step-wise varied situations, e.g., increased previously encoded/decoded elements over modeling steps, we propose a step-adaptive utilization of the three hyper latent representations.
In other words, instead of applying a combined set of the three hyper latent representations, each hyper latent representation is adaptively leveraged at each step.
In addition, we use regional, global, and local information sequentially.
We argue that modeling order is crucial for forward adaptation, which is already known to be a key factor for backward adaptation. 

The green part of \cref{fig:overview} illustrates our proposed contextualization model $c(\cdot)$ at the $i$-th step.
First, the previously modeled $\hat {\bm y}^{< i}$ and the regional feature $\bm \phi_r$ are combined based on the same structure as in the previous approach~\cite{li2023dcvc}, consisting of concatenation, a $1\times1$ convolutional layer, and three DepthConv Blocks.
The DepthConv Block employs depth-wise separable convolutional layers for more efficient implementation.  
Second, the global feature $\bm \phi_g$ is combined with the output of the last DepthConv Block using the Transformer block with cross-attention~\cite{kim2022joint}. 
Finally, we combine the output of the Transformer block with the local feature $\bm \phi_l$ using concatenation followed by three $1\times1$ convolutional layers, yielding the distribution parameters $\bm \mu^i$ and $\bm \sigma^i$.
To make our contextualization model $c(\cdot)$ more efficient in terms of the number of parameters, all layers share weights across the four steps except for the initial $1\times1$ convolutional layer. 

\paragraph{Discussion on modeling order.}
According to the study on theoretical understanding of masked autoencoder via hierarchical latent variable models, the semantic level of the learned representation varies with the masking ratio~\cite{kong2023understanding}. Specifically, extremely large or small masking ratios lead to low-level detailed information such as texture, while non-extreme masking ratios result in high-level semantic information.
Inspired by this, we can infer that the local and global hyper latent representations correspond to relatively low-level information because they are extracted via limited utilization of the latent representation. The receptive field of the local hyper latent representation is limited by 1$\times$1 convolutional layers. While the receptive field of the global hyper latent representation is whole image area, its attention mechanism selectively use the latent representation. Through the same reasoning, we can infer that regional hyper latent representation corresponds to relatively high-level information.
Since different type of contexts has different characteristics, we argue that the modeling order is important for effective entropy modeling.  
In \cref{fig:model_analysis_4}, we empirically confirm that modeling higher-level information (i.e., regional context) first and lower-level information (i.e., local and global contexts) later is more effective than the opposite.
In addition, the order between global and local is shown to be not influential.

\subsection{Adapt}
We design an adaptive entropy model on the quantized latent representation $\hat{\bm y}$ where each element is assumed to follow the Gaussian distribution, and each distribution parameters are obtained from the previous diversification and contextualization stages. 
Following the previous works~\cite{balle2018variational,minnen2018joint}, we formulate our entropy model as follows:
\begin{equation}
p_{\hat{\bm y}}(\hat{\bm y}) = \prod_i \Bigl( \mathcal N\bigl(\mu_i, \sigma_i^2\bigr) \ast \mathcal U\bigl(-\tfrac 1 2, \tfrac 1 2\bigr) \Bigr)(\hat y_i),
\label{eq:entropy-model}
\end{equation}
where $\mu_i$ and $\sigma_i$ are the mean and scale of the Gaussian distribution for each element $\hat y_i$, respectively.

The transforms and entropy model are jointly trained in an end-to-end manner by minimizing the expected length of the bitstream (rate) and the expected distortion between the original image and the decoded image, $d(\cdot, \cdot)$.
When a learned entropy model precisely matches the actual probability distribution, the entropy coding algorithm achieves the minimum rate.
Therefore, we minimize the cross-entropy between the two distributions.
We use mean squared error (MSE) for measuring image distortion. 
The objective function for our method is as follows: 
\begin{equation}
\mathcal{L} = \E_{\bm{x} \sim p_{\bm x}}\bigl[-\log_2 p_{\hat{\bm y}}(\hat{\bm y}) - \log_2 p_{\hat{\bm z}_l}(\hat{\bm z}_l) - \log_2 p_{\hat{\bm z}_r}(\hat{\bm z}_r) - \log_2 p_{\hat{\bm z}_g}(\hat{\bm z}_g) + \lambda \cdot d(\bm x, \hat{\bm x})\bigr],
\end{equation}
where $p_{\bm x}$ is the distribution of the training dataset, the entropy models $p_{\hat{\bm z}_l}$, $p_{\hat{\bm z}_r}$, and $p_{\hat{\bm z}_g}$ are the non-parametric fully factorized entropy models~\cite{balle2017end}, and $\lambda$ is the Lagrange multiplier that determines weighting between rate and distortion.
As the value increases, a model is trained in a direction that reduces information loss, and consequently leads to a higher bit-rate.
\section{Experiments}
\label{sec:experiments}
We use a PyTorch~\cite{pytorch} based open-source library and evaluation platform, CompressAI~\cite{begaint2020compressai}, which has been widely used for developing and evaluating neural image codecs.

\paragraph{Training.}
We set our model parameters as follows: $C=320$, $C_l=10$, $C_r=192$, and $N=8$.
We train our models corresponding six different bit-rates.
We use 300,000 images randomly sampled from the OpenImages~\cite{openimages} dataset.
We construct a batch size of 16 with 256$\times$256 patches randomly cropped from different training images.
All models are trained for 100 epochs using the Adam optimizer.
The learning rate is set to $10^{-4}$ up to 90 epoch, and then decreases to $10^{-5}$.
We use PyTorch v1.9.0, CUDA v11.1, CuDNN v8.0.5, and all experiments are conducted using a single NVIDIA A100 GPU.

\paragraph{Evaluation.}
We evaluate our method on the two popular datasets: Kodak~\citep{kodak} and Tecnick~\cite{tecnick}.
The Kodak dataset consists of 24 images with a resolution of either 768$\times$512 or 512$\times$768 pixels.
The Tecnick dataset is composed of 100 images with a resolution of 1200$\times$1200 pixels.
We evaluate our method in terms of rate--distortion performance.
For this, we calculate the bits per pixel (bpp) after the encoding phase, and measure distortion between the decoded image and the original image using the peak signal-to-noise ratio (PSNR).

\begin{figure}[t]
\centering
\begin{subfigure}{0.48\textwidth}
    \includegraphics[width=\textwidth]{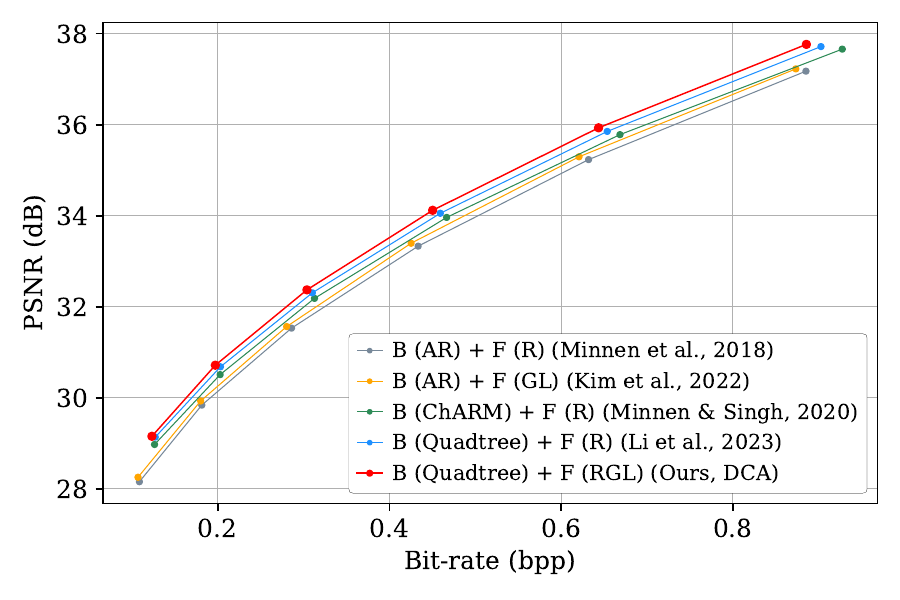}
    \caption{}
    \label{fig:compare_entropy_kodak}
\end{subfigure}
\hfill
\begin{subfigure}{0.48\textwidth}
    \includegraphics[width=\textwidth]{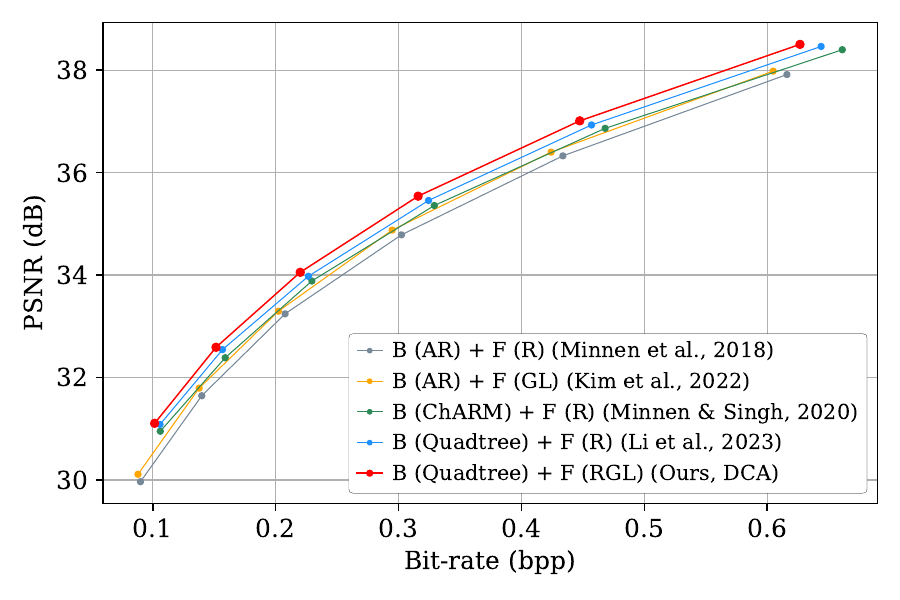}
    \caption{}
    \label{fig:compare_entropy_tecnick}
\end{subfigure}
\caption{Performance comparison with latest entropy models on the two benchmark datasets: (a) Kodak and (b) Tecnick. 
For clear comparisons, we denote each method as follows. B and F mean backward and forward adaptation, respectively, and the corresponding methods are written in parentheses. For backward adaptation, AR, ChARM, and Quadtree represent spatial autoregressive model, channel-wise autoregressive model, and qaudtree partition-based model, respectively. For forward adaptation, L, R, G mean local, regional, and global hyper latent representations, respectively.
}
\end{figure}

\subsection{Comparison with state-of-the-art methods}

We compare the proposed entropy model, DCA, with state-of-the-art entropy models.
DCA can be combined with any transforms; in this paper, DCA is implemented with transforms that have the same structure as in ELIC-sm~\cite{he2022elic}.
For the comparison, we further train image compression methods with four different entropy models~\cite{minnen2018joint,minnen2020channel,kim2022joint,li2023dcvc}.
For a fair comparison, they are also implemented with transforms that have the same structure as in ELIC-sm~\cite{he2022elic}.

\paragraph{Rate--Distortion.}
\cref{fig:compare_entropy_tecnick,fig:compare_entropy_kodak} show the rate--distortion performance on the Kodak and Tecnick datasets, respectively.
The proposed DCA consistently achieves the best rate--distortion performance across all bit-rate regions and two benchmark datasets.
Specifically, DCA achieves 11.96\% average rate savings over VTM-12.1 on the Kodak dataset, while the second (Lie et al. 2023)~\cite{li2023dcvc} and third best (Minnen \& Singh, 2020)~\cite{minnen2020channel} methods obtain 8.55\% and 4.86\%, respectively.

\begin{wrapfigure}[20]{r}{0.5\textwidth}
\vspace{-0.5em}
\centering
\includegraphics[width=\linewidth]{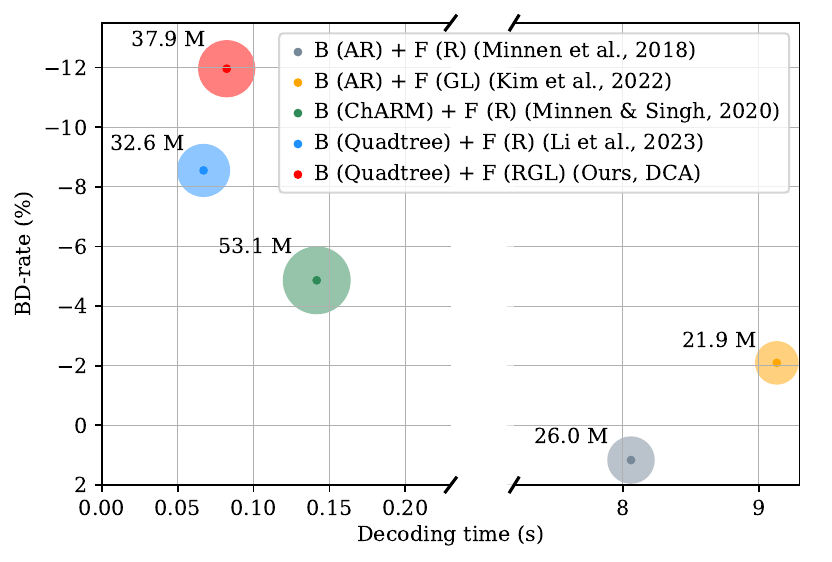}
\caption{Performance comparison with latest entropy models on the Kodak dataset in terms of decoding time, BD-rate, and model size. Decoding time is measured on a NVIDIA V100 GPU. BD-rate means average rate savings over VTM-12.1. The size of the circle is determined proportionally to the number of model parameters, and the specific numbers are written to the left of the circles.}
\label{fig:compare_rdc}
\end{wrapfigure}

\paragraph{Complexity.}
We also evaluate DCA in terms of efficiency.
To this end, we provide the decoding time, the number of model parameters, and Bjøntegaard delta rate (BD-rate)~\cite{bjontegaard2001calculation} in \cref{fig:compare_rdc}.
Decoding time is measured on the Kodak dataset using a single NVIDIA V100 GPU.
BD-rate means the average bit-rate savings compared to a baseline while maintaining the same quality of decoded images.
We set VTM-12.1 as the baseline, calculate BD-rate for each image in the Kodak dataset, and average them.
As shown in \cref{fig:compare_rdc}, our DCA achieves better rate--distortion--computation trade-off than the AR models~\cite{minnen2018joint,kim2022joint} and the ChARM model~\cite{minnen2020channel}.
Even compared to the quadtree-based entropy model~\cite{li2023dcvc}, DCA improves performance significantly, i.e., 3.73\% BD-rate gain, without compromising efficiency as much as possible.

Using the same structure of transforms, we additionally compare DCA with two state-of-the-art entropy models (\cref{tab:rdc}), which shows that DCA improves performance most efficiently. 
It is worth noting that performance improvements are significantly difficult to achieve when there are constraints on compute and memory usage, and this is the achievement of our DCA.

\begin{table}[t]
  \caption{Performance comparison with state-of-the-art entropy models on Kodak.}
  \label{sample-table}
  \centering
  \begin{tabular}{lrrr}
    \toprule
    Methods & BD-rate ($\%$) $\downarrow$ & Decoding time (ms) $\downarrow$ & \# Parameters (M) $\downarrow$ \\
    \midrule
    Baseline (CVPR'23)~\cite{li2023dcvc} & 0.00 & 67.05 & 32.64 \\
    LIC-TCM (CVPR'23)~\cite{liu2023learned} & $-$0.72 & 139.04 & 55.19 \\
    MLIC++ (ICMLW'23)~\cite{jiang2023mlicpp}  & 5.76 & 242.61 & 107.80 \\
    \midrule
    \textbf{DCA (Ours)} & $-$3.73 & 82.05 & 37.89 \\
    \bottomrule
  \end{tabular}
  \label{tab:rdc}
\end{table}

\begin{figure}[h]
\vspace{1em}
\begin{center}
\centerline{\includegraphics[width=\linewidth]{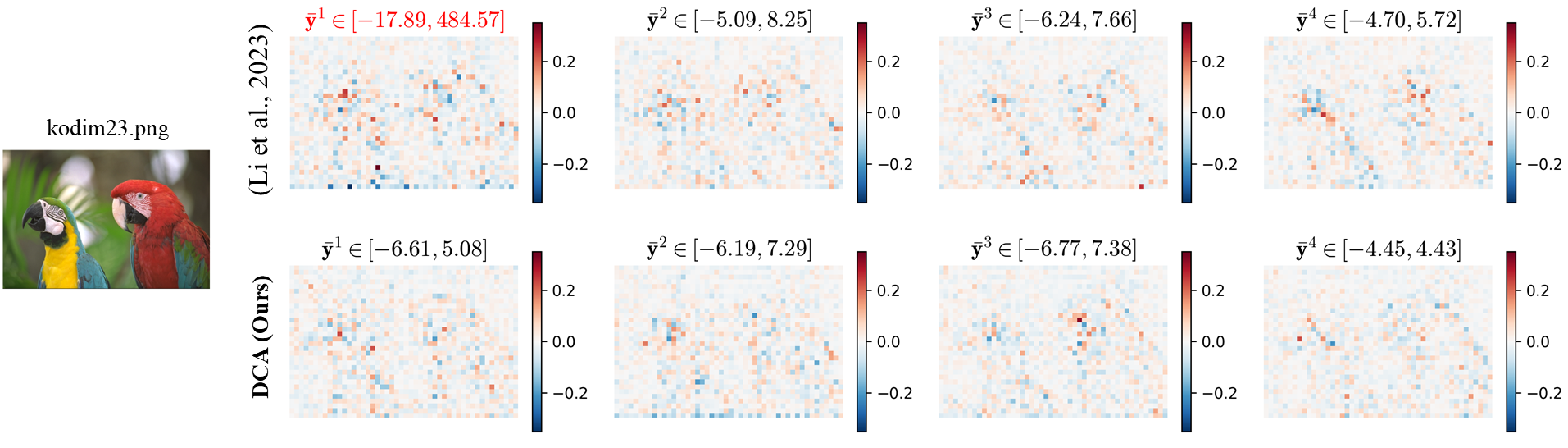}}
\caption{
Illustration of normalized latent representations $\bar{\bm y}^i$ across four steps using both the baseline and proposed DCA. Each sub-figure includes the minimum and maximum values of normalized latent representations. Notably, the baseline exhibits a broader range of values at the first modeling step, resulting in a higher bit-rate. More examples are provided in the appendix.
}
\vspace{-2em}
\label{fig:latent_analysis_1}
\end{center}
\end{figure}

\paragraph{Probability modeling.}
To further show the role of the proposed DCA, we measure the normalized latent representation for each step, i.e., $\bar{\bm y}^i = (\bm{y}^i - \bm{\mu}^i)/\bm{\sigma}^i$.
It provides a standardized measure of how far the latent representation deviates from the predicted mean in terms of estimated standard deviations.
Smaller values indicate that a learned entropy model estimates the true probability distribution more accurately.
In \cref{fig:latent_analysis_1}, we compare the result with that of the baseline model~\cite{li2023dcvc}, which does not leverage diverse contexts for forward adaptation. 
Here, we have an interesting observation: While the existing work shows significantly high values at the first modeling step, DCA demonstrates consistent modeling performance across four steps.
At the first step, since only forward adaptation is possible,
the previous approach can utilize only a limited amount of context.
On the other hand, DCA enables sufficient contextualization through diverse hyper latent representations, addressing the limitation of previous approach, which has difficulty effectively adapting to various situations.

\begin{figure}[h]
\includegraphics[width=\textwidth]{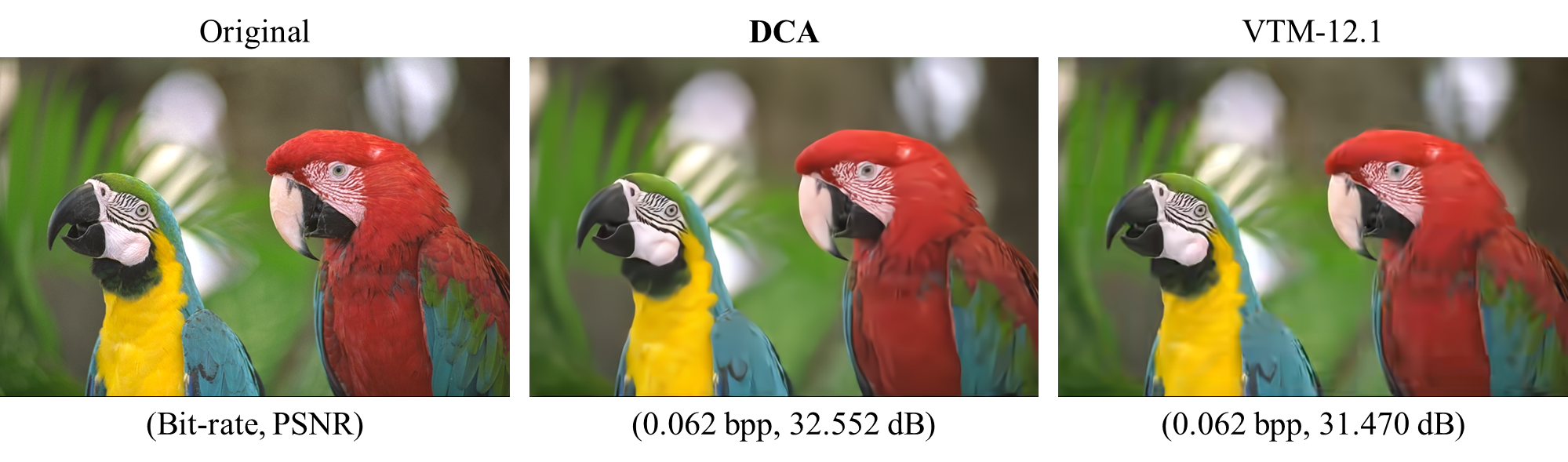}
\caption{Qualitative comparison of the decoded images by the proposed DCA and VTM-12.1.
}
\label{fig:qualitative}
\end{figure}

\paragraph{Qualitative results.}
We provide visual results in \cref{fig:qualitative}, showing the decoded image from DCA has better visual quality and a higher PSNR value than that of VTM-12.1, under the same bit-rate.

\begin{figure}
\begin{minipage}[h]{0.48\textwidth}
\includegraphics[width=\textwidth]{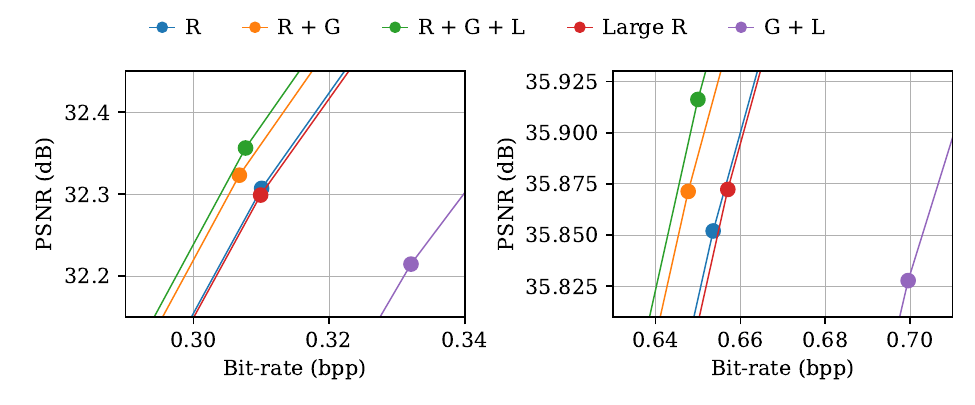}
\captionof{figure}{Analysis of forward context diversity. R, G, L denote the regional, global, and local forward contexts, respectively. ``Large R'' means using larger amount of the regional context.}
\label{fig:model_analysis_1}
\end{minipage}
\hfill
\begin{minipage}[h]{0.48\textwidth}
\includegraphics[width=\textwidth]{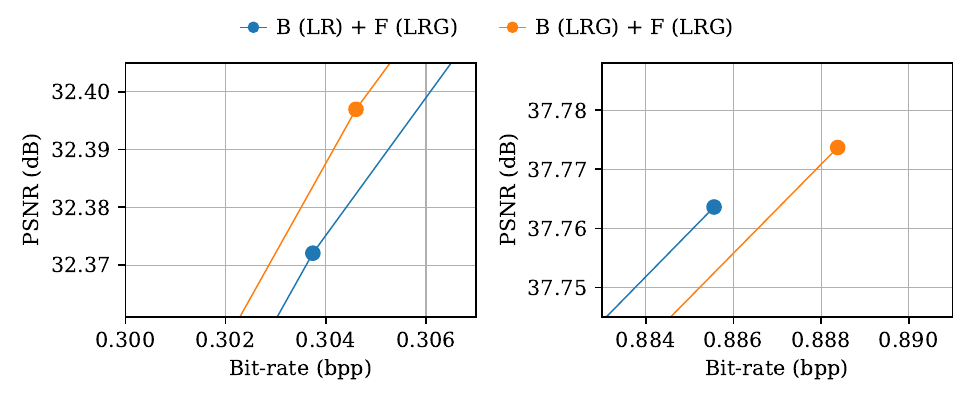}
\captionof{figure}{Analysis of backward context diversity. B and F mean backward and forward contexts, respectively. L, R, G denote the local, regional, and global contexts, respectively.}
\label{fig:model_analysis_5}
\end{minipage}
\end{figure}

\subsection{Model analysis}

\begin{figure}[h]
\begin{minipage}[h]{0.48\textwidth}
\vspace{1em}
\includegraphics[width=\textwidth]{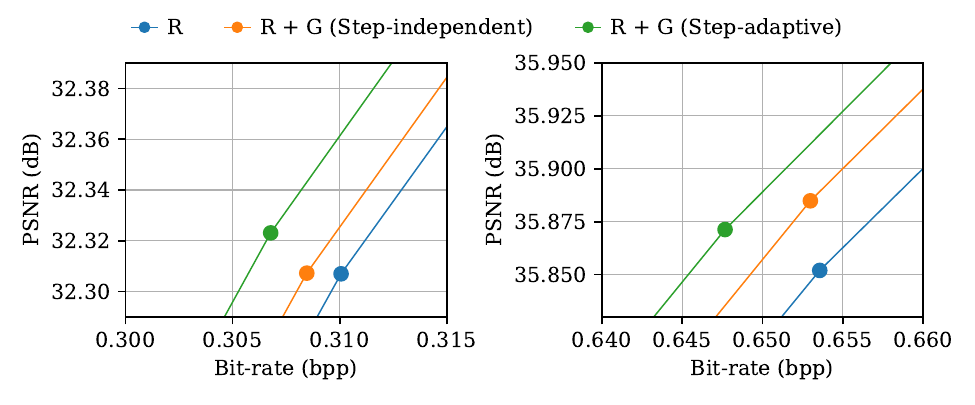}
\captionof{figure}{Analysis of how to contextualize. R and G denote the regional and global contexts, respectively. ``Step-independent'' first combines R and G and then utilizes them for all steps, while ``Step-adaptive'' does not pre-combine them and utilizes each separately for all steps.}
\label{fig:model_analysis_2}
\end{minipage}
\hfill
\begin{minipage}[h]{0.48\textwidth}
\includegraphics[width=\textwidth]{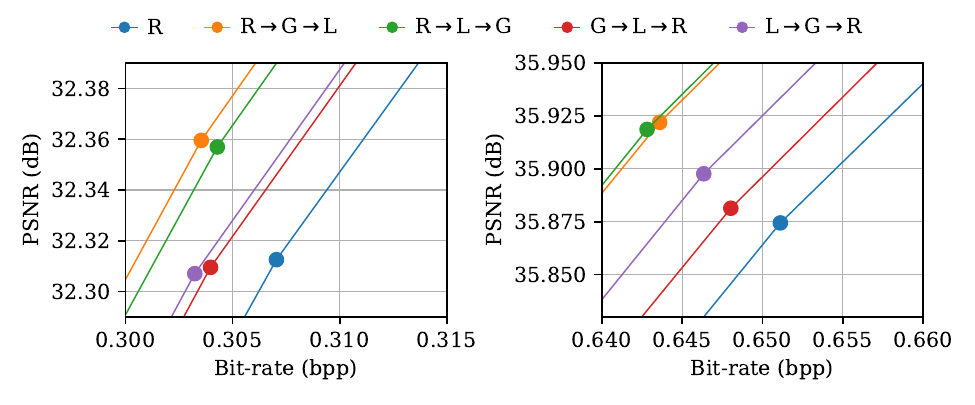}
\captionof{figure}{Analysis of contextualization order. R, G, L denote the regional, global, and local forward contexts, respectively. $\rightarrow$ means the order. For example. the ``R$\rightarrow$G$\rightarrow$L'' method utilizes R, G, and L sequentially.}
\label{fig:model_analysis_4}
\end{minipage}
\end{figure}

\begin{wrapfigure}[12]{r}{0.48\textwidth}
\vspace{-1em}
\centering
\includegraphics[width=\linewidth]{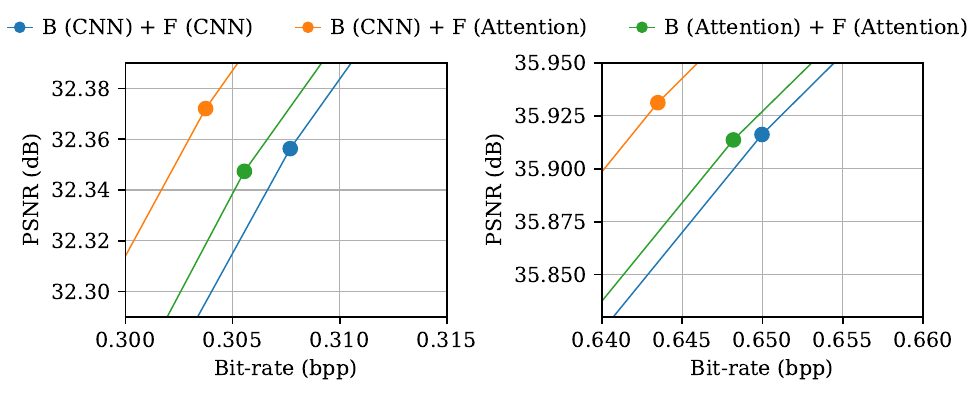}
\caption{Analysis of model architecture for backward and forward adaptation. B and F mean backward and forward adaptation, respectively.}
\label{fig:model_analysis_3}
\end{wrapfigure}

We conduct detailed analyses of DCA. 
To do this, we train different variants of DCA depending on various aspects for analysis.
All results are shown in two different bit-rate regions (\cref{fig:model_analysis_1,fig:model_analysis_2,fig:model_analysis_4,fig:model_analysis_5,fig:model_analysis_3}). 

\paragraph{Analysis of diversification.}
To validate the effectiveness of diversifying contexts for forward adaptation, we compare three different methods depending on the context diversity (\cref{fig:model_analysis_1}): one using regional context (``R''), one using both regional and global contexts (``R + G''), and one using regional, global, and local contexts altogether (``R + G + L'').
We use two additional models: one using a larger regional context (``Large R'') and the other using both global and local contexts (``G + L'').    
The comparison among the first three methods show diversifying forward contexts is effective in both bit-rate regions.
Through the results showing that the ``Large R'' method does not contribute to performance improvement, we once again demonstrate the effectiveness of our diversification.
The last one is decomposing regional context into global and local ones rather than diversifying, which is equal to simply adopting the forward adaptation of Informer~\cite{kim2022joint}.
The result shows the decomposing approach significantly decreases compression efficiency, and diversifying is more effective. 

In addition, while our focus lies on diversifying forward context, someone might be curious about whether the context utilized for the quadtree-based backward adaptation is diverse enough.
To verify this, we conduct experiments additionally extracting global information from the backward context. 
\cref{fig:model_analysis_5}  demonstrates that there is no distinguishable advantage between two: one simply using the quadtree-based method~\cite{li2023dcvc} for backward adaptation (``B (LR) + F (LRG)'') and the other using additional global information for backward adaptation (``B (LRG) + F (LRG)'').
This implies that backward adaptation is favorable when focusing on local and regional contexts.

\paragraph{Analysis of contextualization.}
To show the effectiveness of our contextualization approach, we first compare two different methods in \cref{fig:model_analysis_2}.
``R + G (Step-independent)'' combines regional and global information in advance and utilizes the combined one regardless of the step. 
The other adaptively utilizes regional and global information separately for each step, ``R + G (Step-adaptive)''.
As a reference, we use the model utilizing only regional context, i.e., ``R''. 
The result shows that both are effective and our step-adaptive approach is more beneficial for boosting performance.

In addition, we analyze the effectiveness of our modeling order for contextualization in \cref{fig:model_analysis_4}.
Four different ordering methods and one reference method are used for the comparison.
Our ordering approach (``R$\rightarrow$G$\rightarrow$L'') achieves the best rate--distortion performance, and the methods are categorized into two groups based on the performance in both bit-rate regions. 
Models that prioritize the regional context (``R$\rightarrow$G$\rightarrow$L'' and ``R$\rightarrow$L$\rightarrow$G'') show better performance compared to those that do not prioritize it (``G$\rightarrow$L$\rightarrow$R'' and ``L$\rightarrow$G$\rightarrow$R'').
We observe that the method using the opposite modeling order of the proposed sequence (``L$\rightarrow$G$\rightarrow$R'') even exhibits a performance decline compared to the baseline (``R'') in a lower bit-rate region.

\paragraph{Analysis of architecture.}
Applying attention mechanisms on various tasks is one of the most actively researched topics.
We analyze the effect of the architecture combination for forward and backward adaptation in DCA.
\cref{fig:model_analysis_3} compares three different methods: one without attention (``B (CNN) + F (CNN)''), another applying attention only to forward adaptation (``B (CNN) + F (Attnention)''), and the last applying attention for both (``B (Attention) + F (Attention)'').  
The results show that CNN and attention are effective for forward and backward adaptation, respectively.
We infer that focusing on local and regional information is preferable in backward adaptation; thus, a CNN with a locality inductive bias may be more effective.
\section{Conclusion}
\label{sec:conclusion}

In this paper, we proposed a fast and effective entropy modeling framework, DCA, which diversifies forward contexts by extracting local, regional, and global information, and contextualizes current elements with the diverse forward and backward contexts.
We demonstrated that our DCA improves rate--distortion performance significantly compared to previous approach without compromising efficiency as much as possible.
Furthermore, we provided diverse insights into entropy modeling by conducting a comprehensive and in-depth analysis of the design aspects of DCA.

\paragraph{Limitation and future works.}
To address the limitation of the state-of-the-art entropy models, we focused on paving a novel framework with diverse contexts rather than designing neural architectures.
Therefore, DCA can be limited by the architectural designs that are inspired by the existing works~\cite{kim2022joint,zhu2022transformer,li2023dcvc}.
In the future, we expect that it would be further improved by neural architectures especially designed for the diverse contexts. In addition, it is worth exploring alternative criteria for diversification beyond the spatial range the contexts covers (i.e., local, regional, and global contexts).


{
\small
\bibliographystyle{abbrvnat}
\bibliography{neurips_2024}
}

\newpage
\appendix

\section{Additional results}

\begin{figure}[!hb]
\begin{center}
\centerline{\includegraphics[width=0.96\linewidth]{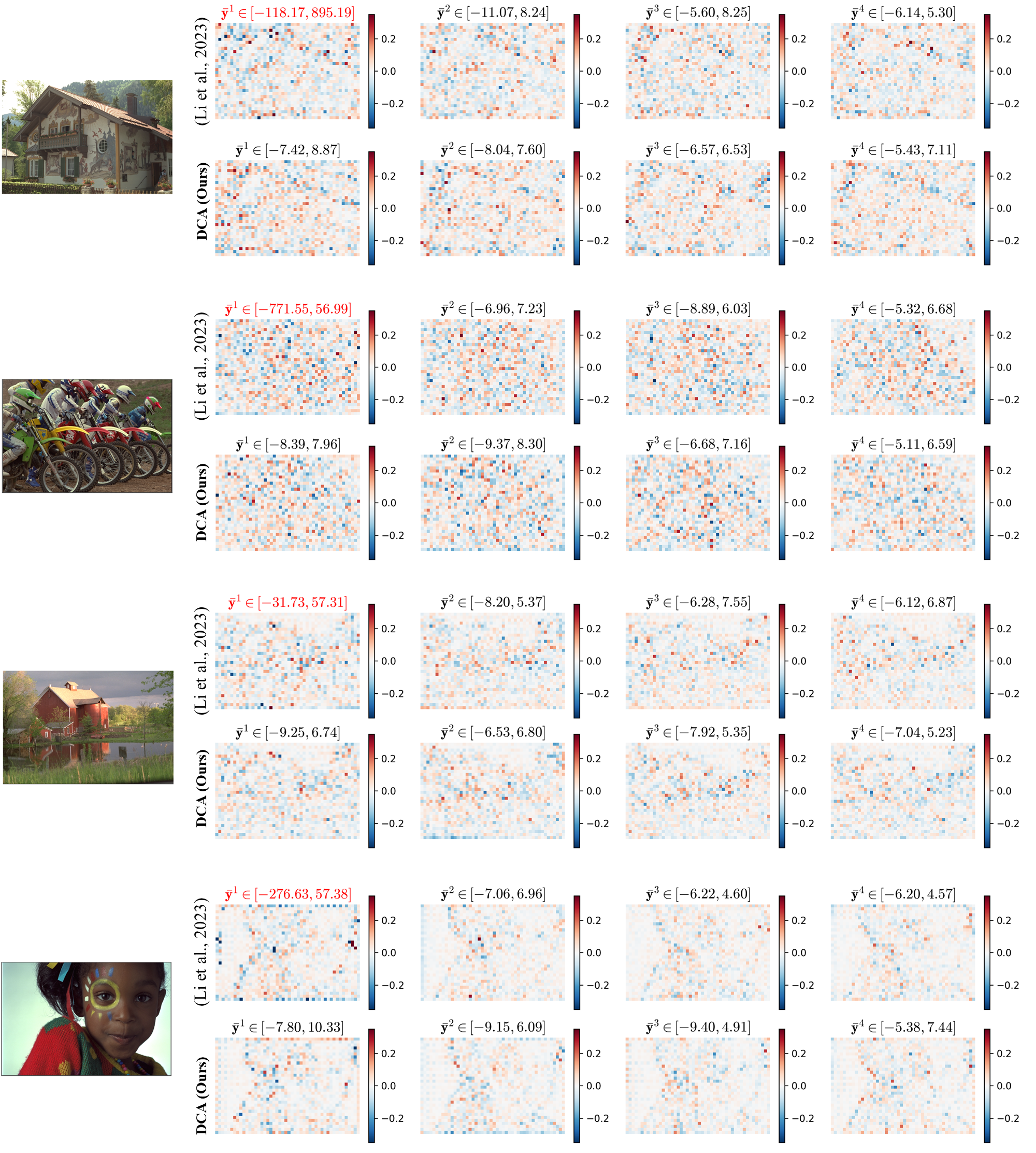}}
\caption{
Illustrations of normalized latent representations $\bar{\bm y}^i$ across four steps for Kodak images using both the baseline~\cite{li2023dcvc} and proposed DCA. Each sub-figure includes the minimum and maximum values of normalized latent representations. Notably, the baseline exhibits a broader range of values at the first modeling step, resulting in a higher bit-rate.
}
\label{fig:latent_analysis_3}
\end{center}
\vspace{-1em}
\end{figure}

We provide the normalized latent representations of images from the Kodak dataset in \cref{fig:latent_analysis_3}.

We provide an in-depth runtime analysis by sub-systems in \cref{tab: runtime}.

\begin{table}[h]
  \caption{Runtime (ms) of DCA. Total encoding/decoding time includes the ANS entropy coding.}
  \label{tab: runtime}
  \centering
  \begin{tabular}{ccccccccccc}
    \toprule

    \multicolumn{2}{c}{Transforms} & \multicolumn{7}{c}{Entropy model (DCA)} & \multicolumn{2}{c}{Total} \\
    \cmidrule(r){1-2} \cmidrule(r){3-9} \cmidrule(r){10-11}
    
    $f_a$ & $f_s$ & $l_a$ & $l_s$ & $r_a$  & $r_s$ & $g_a$ & $g_s$ & $c$ & Encoding & Decoding \\
    \midrule 
    3.46 & 1.54 & 0.98 & 0.96 & 5.32 & 4.78 & 0.63 & 0.08 & 14.75 & 117.49 & 82.05 \\

    \bottomrule
  \end{tabular}
\end{table}

\end{document}